\documentclass[journal]{IEEEtran}

\ifCLASSINFOpdf
  \usepackage[pdftex]{graphicx}
  \graphicspath{{./figs/}}
  \DeclareGraphicsExtensions{.pdf,.jpeg,.png}
\else
\fi
\usepackage[cmex10]{amsmath}
\usepackage{mathtools}
\usepackage{amssymb}
\usepackage{algorithm}
\usepackage{algpseudocode}
\usepackage[table,xcdraw]{xcolor}

\usepackage{array}

\usepackage{booktabs}
\usepackage{multirow}
\usepackage{tabu}

\usepackage{url}
\PassOptionsToPackage{hyphens}{url}\usepackage{hyperref}

\usepackage{biograph}        
\usepackage{afterpage}

\usepackage{cite}
\usepackage{multirow} 
\usepackage{rotating}

\usepackage{float}
\usepackage{color}
\usepackage{soul}

\usepackage{colortbl}

\def\por1{\partial}

\newcolumntype{S}{>{\centering\arraybackslash} m{.4\linewidth} }

\makeatletter
  \newcommand\tinyv{\@setfontsize\tinyv{5pt}{7}}
\makeatother

\newlength{\hspacephantom}
\begin{document}
\DeclareGraphicsExtensions{.pdf,.jpeg,.png}

\title{Efficient No-Reference Quality Assessment and Classification Model for Contrast Distorted Images}




\author{Hossein Ziaei Nafchi, \textnormal{and} Mohamed Cheriet, \IEEEmembership{Senior Member,~IEEE} 
\thanks{Copyright (c) 2018 IEEE. Personal use of this material is permitted. However, permission to use this material for any other purposes must be obtained from the IEEE by sending a request to pubs-permissions@ieee.org.}
\thanks{H. Ziaei Nafchi, and M. Cheriet are with the Synchromedia Laboratory for Multimedia Communication in Telepresence,
\'Ecole de technologie sup\'erieure, Montreal (QC), Canada H3C 1K3 (email: hossein.zi@synchromedia.ca; mohamed.cheriet@etsmtl.ca}
\thanks{Manuscript received ? ?, ?; revised ? ?, ?.}}

\markboth{}%
{}
%


\maketitle

\begin{abstract}
In this paper, an efficient Minkowski Distance based Metric (MDM) for no-reference (NR) quality assessment of contrast distorted images is proposed. It is shown that higher orders of Minkowski distance and entropy provide accurate quality prediction for the contrast distorted images. The proposed metric performs predictions by extracting only three features from the distorted images followed by a regression analysis. Furthermore, the proposed features are able to classify type of the contrast distorted images with a high accuracy. Experimental results on four datasets CSIQ, TID2013, CCID2014, and SIQAD show that the proposed metric with a very low complexity provides better quality predictions than the state-of-the-art NR metrics. The MATLAB source code of the proposed metric is available to public at http://www.synchromedia.ca/system/files/MDM.zip. 
\end{abstract}


%
\IEEEpeerreviewmaketitle


\maketitle

\section{Introduction}
\label{sec:intro}


\IEEEPARstart{I}{mage} quality assessment (IQA) is a very important step in many image processing applications such as monitoring, benchmarking, restoration and parameter optimization \cite{SSIM}. Human visual system can easily have a fair judgment on the quality of the images. However, subjective assessment of images is a very time consuming task. Hence, many IQA models (IQAs) have been proposed to automatically provide objective quality assessment of images \cite{SSIM, MAD, VIF, FSIM, GMSD, VSI, Metrics2011, NRwavelet, NRdct, NRspatial, NRgradient, NIQE, NRCVPR}. Among them, NR-IQAs \cite{NRwavelet, NRdct, NRspatial, NRgradient, NIQE, NRCVPR, NFERM} are of high interest because in most present and emerging practical real-world applications, the reference signals are not available \cite{SPM2011}. NR-IQAs do not need any information on the reference image. It is worth to mention that reduced-reference (RR) metrics \cite{RR2005, RRaware, RR2009} need partial information about the reference image and that the full-reference (FR) metrics \cite{SSIM, MAD, VIF, FSIM, GMSD, Metrics2011, MDSI, PSIM} require the reference image.

\begin{figure}[htb]
\scriptsize
\begin{minipage}[b]{0.32\linewidth}
  \centering
  \centerline{\includegraphics[height=1.8cm]{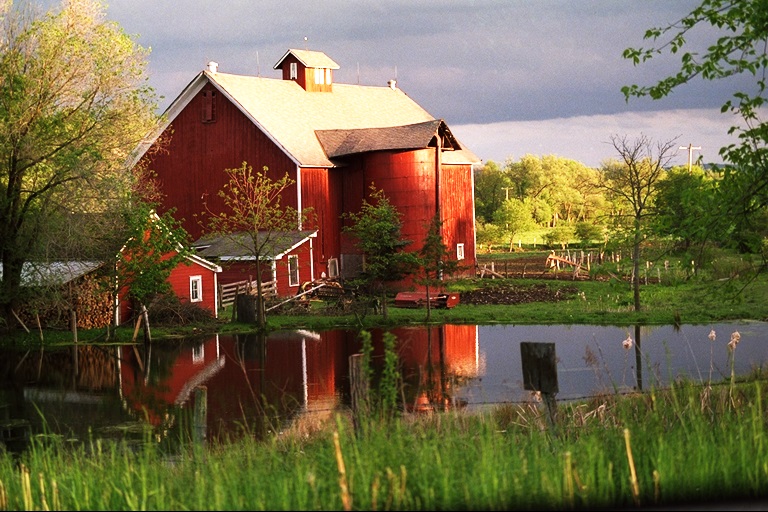}}
\end{minipage}
\begin{minipage}[b]{.32\linewidth}
  \centering
  \centerline{\includegraphics[height=1.8cm]{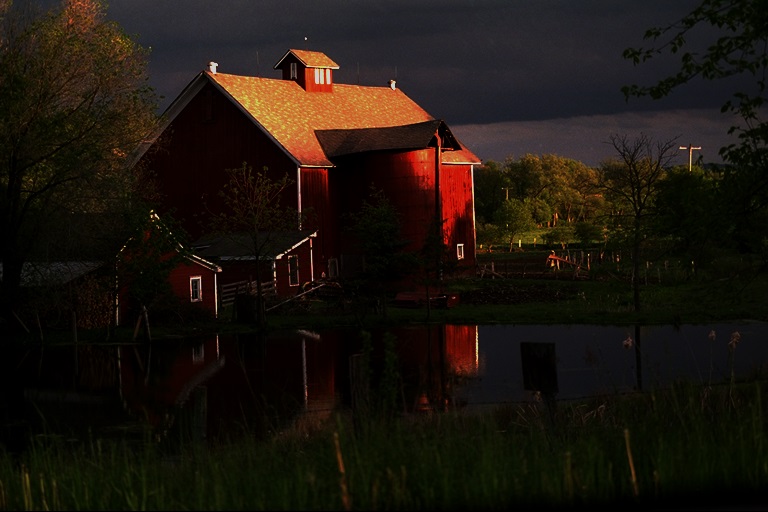}}
\end{minipage}
\begin{minipage}[b]{0.32\linewidth}
  \centering
  \centerline{\includegraphics[height=1.8cm]{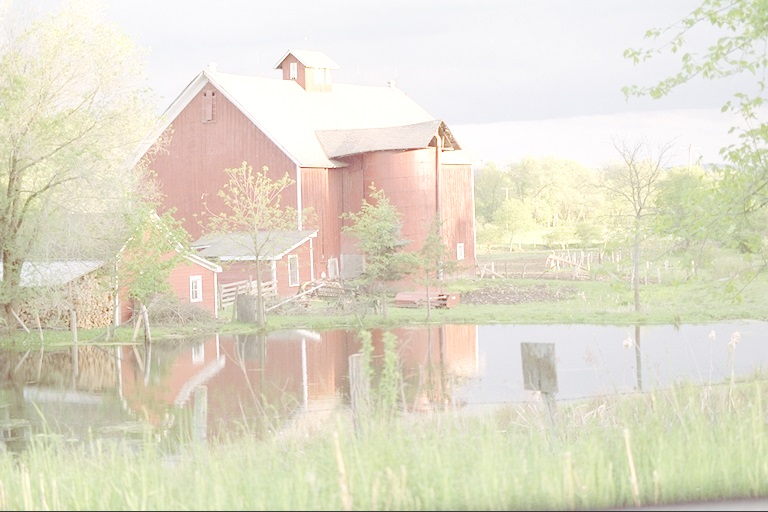}}
\end{minipage}

\vspace{0.2cm}

\begin{minipage}[b]{0.32\linewidth}
  \centering
  \centerline{\includegraphics[height=1.8cm]{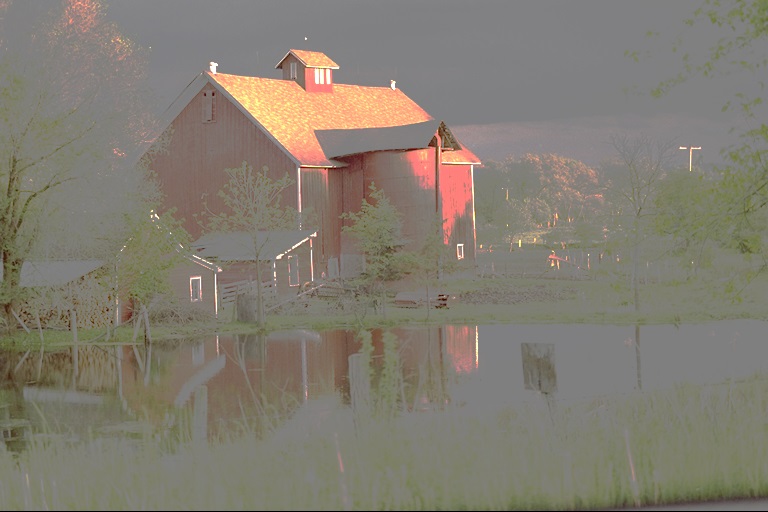}}
\end{minipage}
\begin{minipage}[b]{0.32\linewidth}
  \centering
  \centerline{\includegraphics[height=1.8cm]{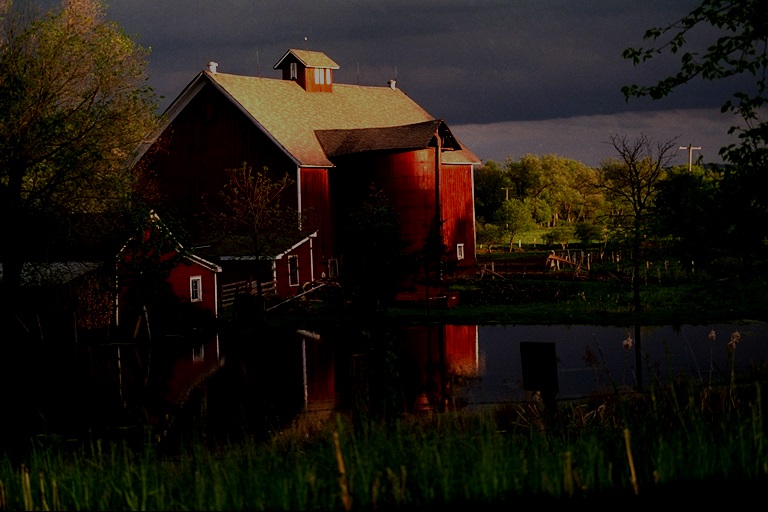}}
\end{minipage}
\begin{minipage}[b]{0.32\linewidth}
  \centering
  \centerline{\includegraphics[height=1.8cm]{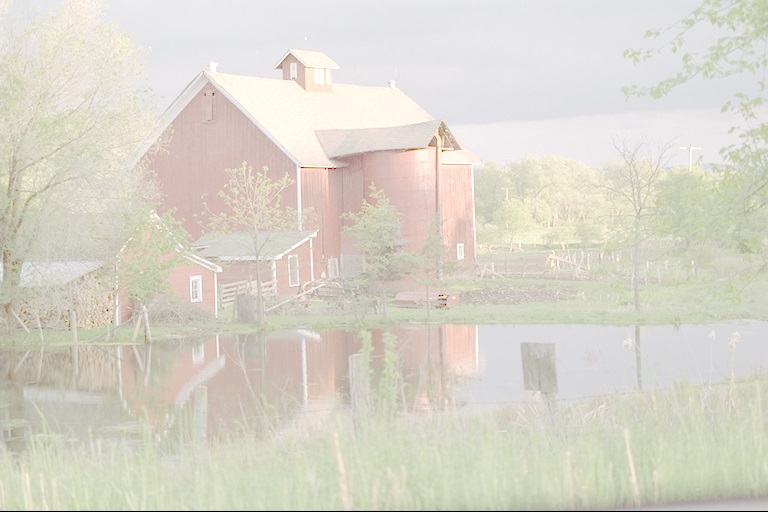}}
\end{minipage}
\caption{Sample contrast distorted images from CCID2014 \cite{cyber2015}.}
\label{fig1}
\end{figure}

Contrast distortion, which lies within the scope of this paper, is commonly produced in image acquisition setup. Poor and varying illumination conditions and poor camera's quality can drastically change image contrast and visibility. Fig. \ref{fig1} shows the six examples of contrast distorted images. Several contrast enhancement methodologies have been proposed to adjust image contrast. These methods may over/under estimate the amount of contrast distortion and fail at enhancement accordingly. These methods, however, can use prior information provided by IQAs to overcome this wrong estimation.

With introduction of quality aware images \cite{RRaware}, RR-IQAs have shown their usefulness at assessment of image distortions caused by transmission in particular. Prior information about reference image is embedded inside the image to be transmitted, and receiver decodes this information and uses it for quality assessment and even correction of distortions. The resulting metrics that eventually doesn't need training are good examples to illustrate RR-IQAs. In \cite{RRICIP2013}, a RR-IQA called the RIQMC was proposed to assess the quality of contrast distorted images. RIMQC is a two-step model that uses entropy and four order statistics, e.g. mean, standard deviation, skewness and kurtosis. These are then linearly combined and a quality score is calculated. Seven parameters of the RIQMC are trained based on the 322 images of the CID2013 dataset that also introduced in \cite{RRICIP2013}. The performance of the RIQMC is very high and at the level of the leading FR-IQA models. The RIQMC was further modified in \cite{cyber2015} by computing the phase congruency of the reference and distorted images. In \cite{QMC}, a more efficient RR-IQA called QMC was proposed that uses entropy and saliency features of the reference and distorted images for quality prediction. RCIQM is a more recent RR-IQA model that benefits from a bottom-up and top-down strategy \cite{RCIQM2017}. It is based on bottom-up analysis of the free energy principle and top-down analysis of histograms of the reference and distorted images. RCIQM delivers a high performance for quality assessment of the contrast distorted images. The problem with these RR-IQAs is that they necessarily need reference or original image to be available.    

There are limited methods in order to assess quality of the contrast distorted images \cite{SPL2014, NIQMC2017}. The authors in \cite{SPL2014} use a natural scene statistics (NSS) induced model to blindly predict the quality of contrast distorted images. They also use five features based on the NSS models of mean, standard deviation, skewness, kurtosis and entropy. Then, support vector regression is utilized to find a mapping function between these five feature set and subjective quality scores. They used 16873 images to train their NSS model. The NR-IQA model in \cite{NIQMC2017} called NIQMC takes into account both local and global aspects of the contrast distorted images. In the local part, entropy of salient regions is computed. For the global part, a histogram analysis is proposed. NIQMC provides accurate quality predictions for contrast distorted images. The problem with this method is its high computational time. Four features including the entropy of the phase congruency edge strength and three contrast energy-based features are used by BIQME \cite{BIQME} for contrast distortion assessment. BIQME is a general-purpose opinion unaware NR-IQA model with total number of 17 features and shows high performance, specially for the contrast distorted images.

In this paper, we propose a NR-IQA metric that is highly efficient and provides high prediction accuracy at the same time. We have found that the standard deviation (SD) alone provides a moderate quality prediction accuracy for global contrast distorted images. The SD when used to compare contrast level of two images is called root-mean-square (rms) contrast \cite{rms1990}. The promising performance of rms contrast for global contrast changed images motivates us to use a variation of the Minkowski distance formulation along with the power-law transformation for no-reference quality assessment of contrast distorted images (NR-CDIQA). Power-law transformations are traditional image processing techniques that have been previously used for gamma correction and contrast manipulation. Previously, the Minkowski distance has been mainly used in IQA for two main purposes. The Minkowski metric has been used as a FR-IQA metric \cite{handbook2000}, and the Minkowski pooling as a pooling strategy \cite{Mpooling2006}. Minkowski error metric between reference image $\mathcal{R}$ and distorted image $\mathcal{D}$  is defined as:

\begin{equation}
  \ \text{E}_\rho = \Big(\sum_{i=1}^{N}\big| \mathcal{R}_i - \mathcal{D}_i \big|^\rho \Big)^{1/\rho}.
  \label{equ:minkowski1}
\end{equation}          
where $N$ is the number of image pixels, and $\rho\geq1$ refers to the Minkowski power. Also, given any local similarity ($S$) map computed between a reference and distorted image by an IQA model, the Minkowski pooling is defined as

\begin{equation}
  \ \text{M} = \frac{1}{N}\sum_{i=1}^{N}S_i^\rho.
  \label{equ:pooling}
\end{equation}                    
where M is the quality score of that IQA model. Except for the case $\rho=1$ which is equal to the mean pooling, Minkowski pooling is rarely used in the literature \cite{GMSD, MDSI}. 

In this paper, we use higher orders of the Minkowski distance along with the power-law transformation and entropy to provide accurate quality predictions for contrast distorted images. In addition, the features of the proposed metric are able to classify type of the contrast distorted images. This information can be very useful in enhancing the contrast distorted images in real-time. To the best of our knowledge, classification of contrast distortion types has not been considered in the literature. In the following, the main contributions of the paper as well as its differences with respect to the previous works are briefly explained.

The proposed metric uses higher orders of Minkowski distance along with the power-law transformations, while in previous works like \cite{rms1990, SPL2014}, only the rms contrast or second order image statistic is used. To the best of our knowledge, Minkowski distance has not been used for the purpose of no-reference image quality assessment.      

Entropy is widely used in previous studies for the purpose of contrast distortion assessment \cite{RRICIP2013, SPL2014, cyber2015, QMC, RCIQM2017, NIQMC2017}. The proposed NR-IQA metric also uses entropy but despite having much lower complexity delivers higher and more consistent predictions than existing NR-IQA models on different datasets.

The three features of the proposed method are able to classify the type of contrast distorted images with a high accuracy, while features of existing method are not suitable for this task.

\section{Proposed Metric (MDM)}
\label{method}

\begin{figure}[htb]
\begin{minipage}[b]{0.99\linewidth}
  \centering
  \centerline{\includegraphics[height=5.8cm]{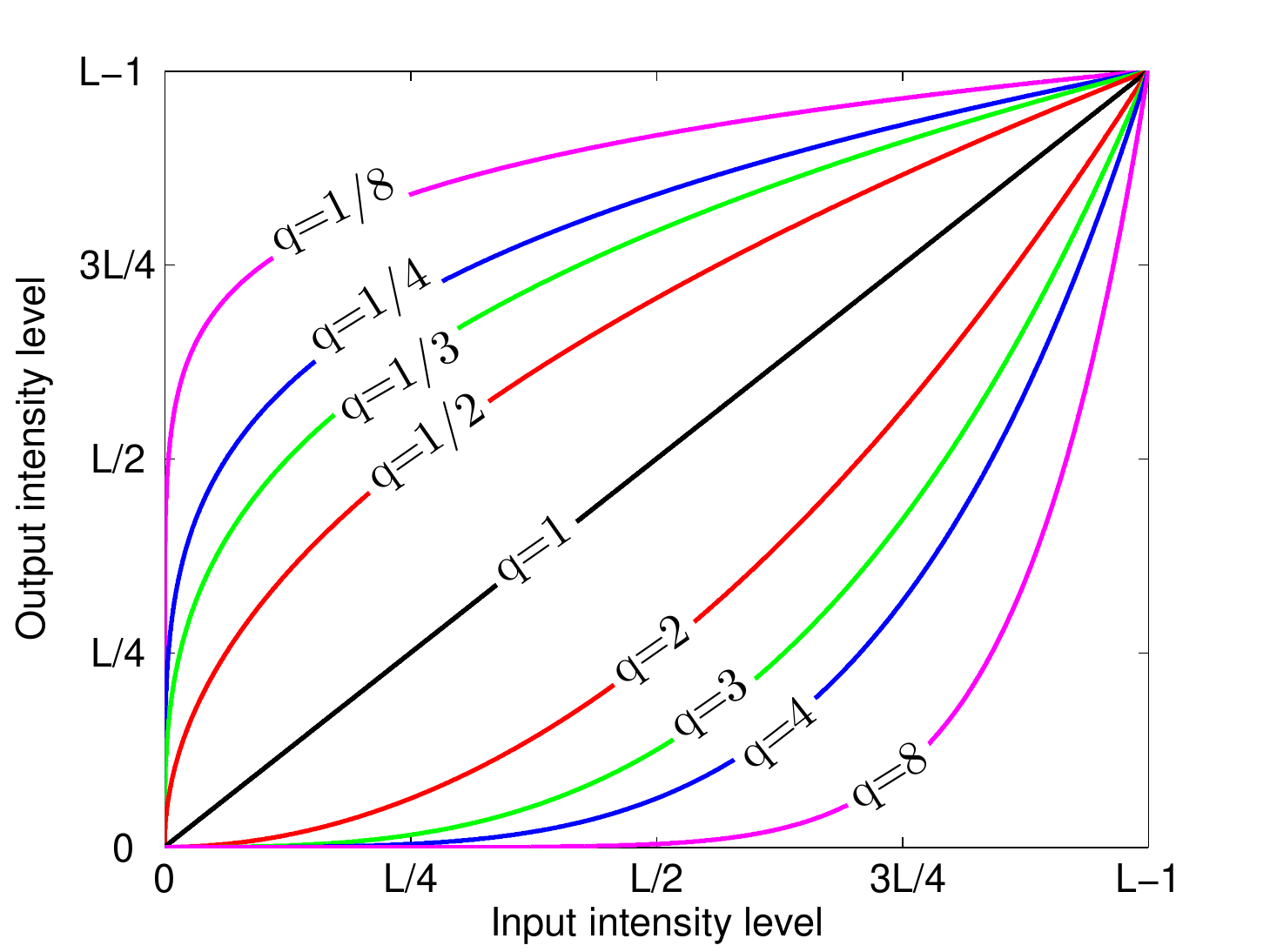}}
\end{minipage}
\caption{Output intensity level versus input intensity level ($q=1$) for different values of $q$.}
\label{gamma}
\end{figure}

The proposed NR-IQA of contrast distorted images follows the Minkowski distance formulation. Let's define the deviation as the variation of data values compared to a measure of central tendency (MCT) such as the mean, median, or mode. A deviation is in fact Minkowski distance of order $\rho$ between an arbitrary vector \textbf{x} and its MCT:  

\begin{equation}
  \ \text{D}(\textbf{x},\rho) = \Big(\sum_{i=1}^{N}\big| \textbf{x}_i - \text{MCT(\textbf{x})} \big|^\rho \Big)^{1/\rho}.
  \label{equ:minkowski2}
\end{equation}                    
where, $\textbf{x}_i$ denotes a vector value, MCT refers to the mean value of vector \textbf{x}, and $\rho \geq 1$ indicates to the type of deviation. The proposed NR-IQA model uses a variation of the equation (\ref{equ:minkowski2}) as follows:

\begin{equation}
  \ \widehat{\text{D}}(\textbf{x},\rho) = \Big( \frac{1}{N} \sum_{i=1}^{N}\big| \textbf{x}_i - \text{MCT(\textbf{x})} \big|^\rho \Big)^{1/\rho}.
  \label{equ:minkowski3}
\end{equation}                    
where, $N$ is the number of pixels in an image and $\frac{1}{N}$ accounts for image resolution. Equation (\ref{equ:minkowski3}) is equivalent to the mean absolute deviation for $\rho=1$ and equivalent to the standard deviation (rms contrast) for $\rho=2$. Let $\mathcal{D}$ denote the distorted image and $\mathcal{D}^q$ denote the pixel-wise distorted image $\mathcal{D}$ to the power $q$, which is known as power-law transformation. Also let MCT$^q$ denote the mean value of the image $\mathcal{D}^q$. The proposed NR metric MDM for distorted image $\mathcal{D}$ is computed by the following equation:

\begin{equation}
  \ \text{MDM}^{\rho, q}(\mathcal{D}) = \sqrt[4]{ \Big( \frac{1}{N} \sum_{i=1}^{N}\big| \mathcal{D}^q_i - \text{MCT}(\mathcal{D}^q) \big|^\rho \Big)^{1/\rho} }.
  \label{equ:MDM1}
\end{equation}

where, $\mathcal{D}_i^q$ denotes one pixel of distorted image to the power of $q$. The fourth root in above equation is used for better numerical stability and visualization of quality scores. The reason for inclusion of parameter $q$ is that contrast distorted images may follow the gamma transfer function in the form of $\mathcal{D}$ = $\mathcal{R}^q$. In this paper, a large value of $q$ is used. This large value of $q$ will most likely increase the severity of distortion of $\mathcal{D}$. From one perspective, this effect can be compared with the strategy proposed in \cite{blur2007}. In \cite{blur2007}, the input image is blurred and the result is compared with the input image in order to blindly assess its blurriness. Here, a similar strategy is used except that $\mathcal{D}^q$ is not compared with $\mathcal{D}$. Fig. \ref{gamma} illustrates the impact of parameter $q$ on the input intensity level.

Equation \ref{equ:MDM1} computes the first feature of the proposed metric for a distorted image $\mathcal{D}$. The proposed metric also computes a second feature by the same equation from the complement of a contrast distorted image, e.g. $\overline{\mathcal{D}} = 255 - \mathcal{D}$. Except for some special cases, $\text{MDM}^{\rho, q}(\mathcal{D}) \neq \text{MDM}^{\rho, q}(\overline{\mathcal{D}})$. Note that rms contrast of $\mathcal{D}$ and $\mathcal{\overline{D}}$ are equal. In experiments, the values of $\rho$ and $q$ are set to 64 and 8.

\begin{figure}[htb]
\begin{minipage}[b]{0.99\linewidth}
  \centering
  \centerline{\includegraphics[height=5.8cm]{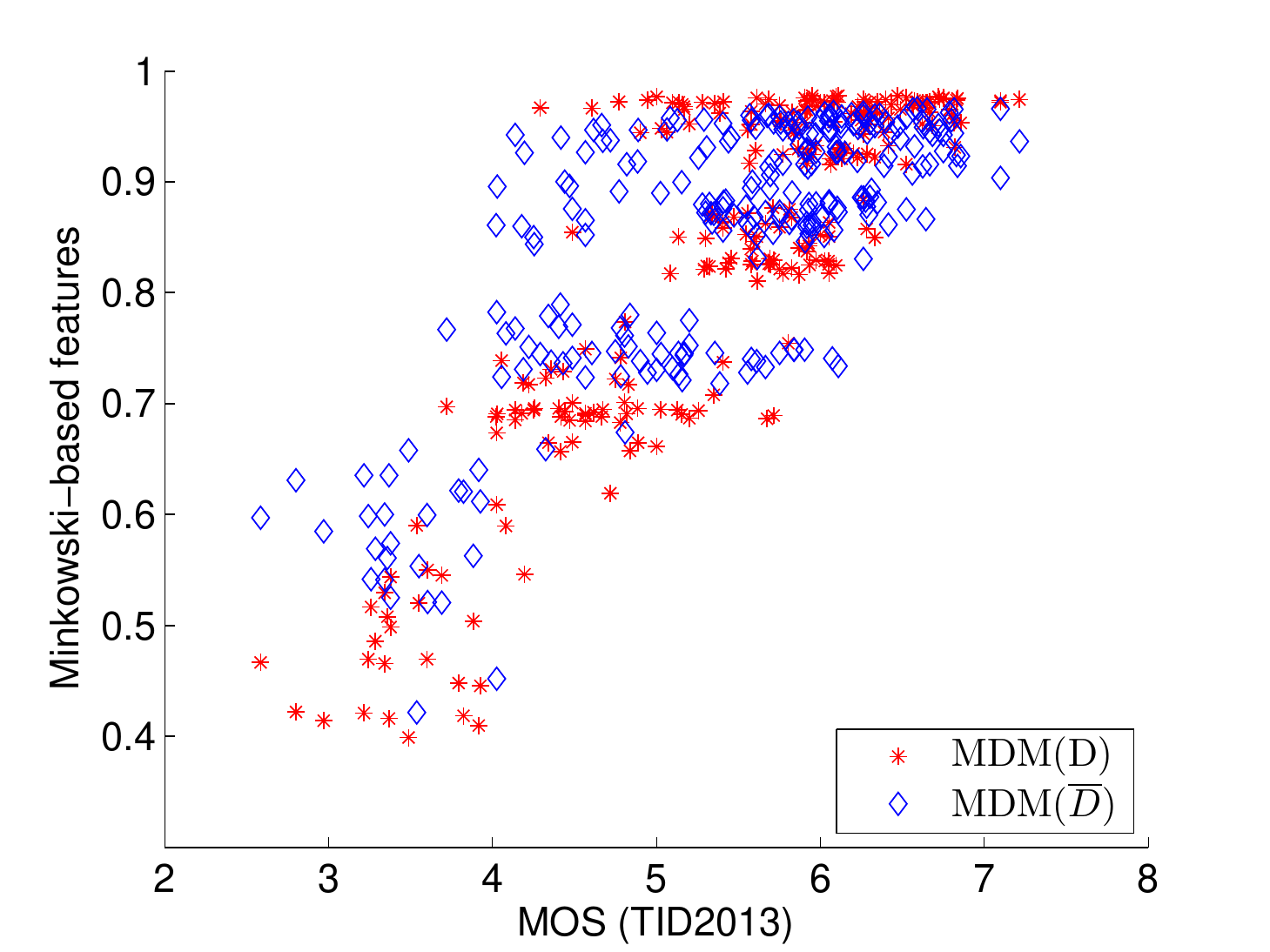}}
\end{minipage}
\caption{Two Minkowski-based features versus MOS for TID2013 dataset.}
\label{MDMonTID2013}
\end{figure}

The two MDM based features are highly suitable for quality assessment of global contrast change ($\mathcal{D} = \mathcal{R}^q$) and mean shift ($\mathcal{D} = \mathcal{R} \pm \bigtriangleup $) distorted images, where $\bigtriangleup$ is a scalar within dynamic range of $\mathcal{R}$. Fig. \ref{MDMonTID2013} shows values of these two features versus MOS values for 250 contrast distorted images of TID2013 dataset \cite{TID2013}. This plot shows that values of the proposed Minkowski-based features are proportional to the MOS values.

\begin{figure}[htb]
\begin{minipage}[b]{0.99\linewidth}
  \centering
  \centerline{\includegraphics[height=5.8cm]{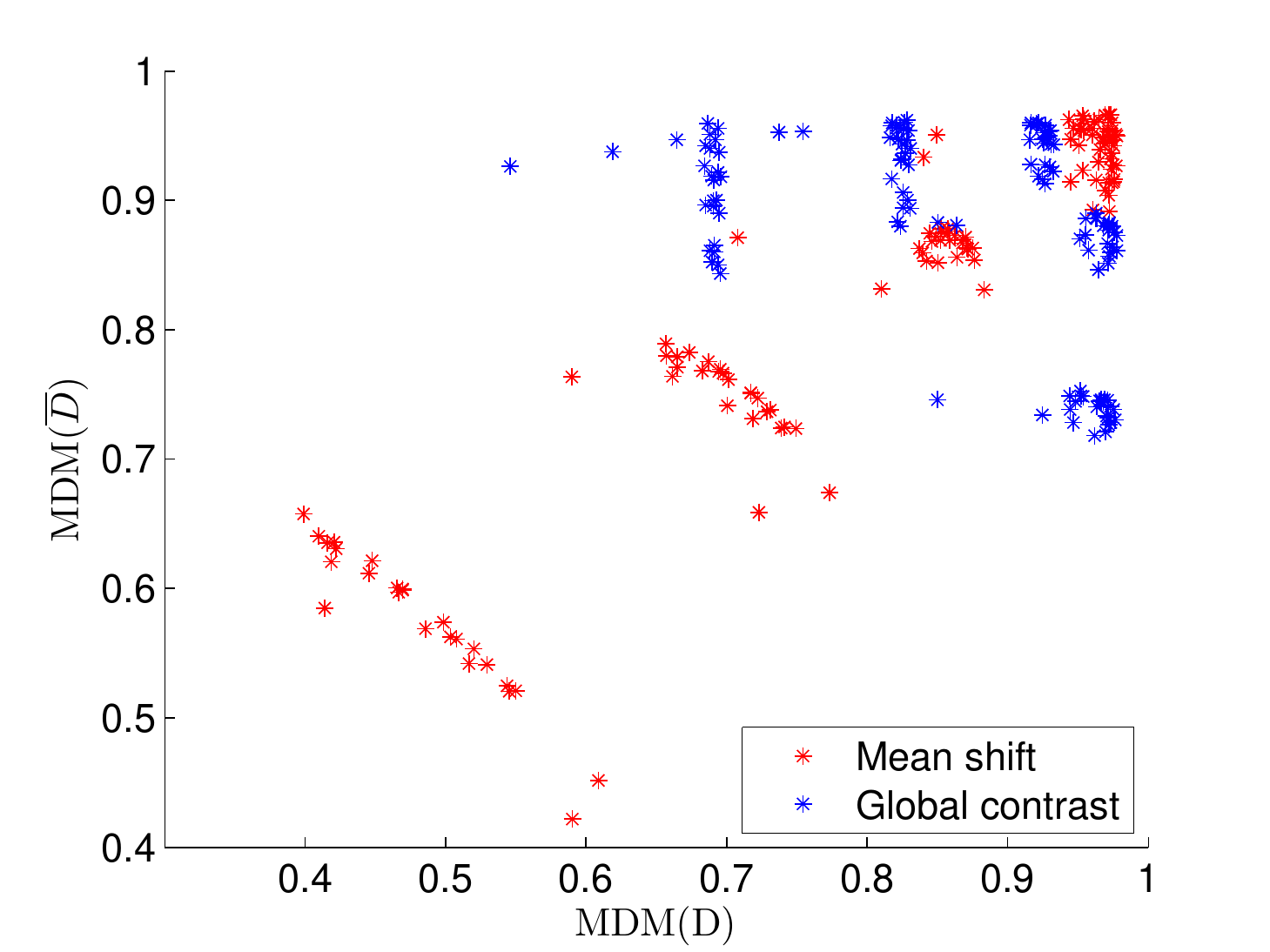}}
\end{minipage}
\caption{Visualization of the two Minkowski based features $\text{MDM(D)}$ and MDM($\overline{D}$) for two contrast distortion types.}
\label{MDMtwoCLASS}
\end{figure}

In addition, the proposed Minkowski-based features can be used to classify contrast distortion types. Fig. \ref{MDMtwoCLASS} shows plot of the first versus second Minkowski-based feature. The points of each distortion type on the plot can be separated with a high accuracy which shows the ability of these two features for classifying contrast distortions.

Additionally, the proposed metric uses entropy of the distorted image along with the MDM features as the third feature. Entropy is a common statistical measure of randomness which is useful in analyzing texture of the images. Previous study \cite{cyber2015} states that high-contrast image often has large entropy. The entropy is defined as:

\begin{equation}
  \ \text{H}(\mathcal{D}) = -\sum_{L=0}^{255} \text{P}_L(\mathcal{D})\log_2 \text{P}_L (\mathcal{D})
  \label{equ:entropy}
\end{equation}                    
where, P$_L(\mathcal{D})$ is the probability density of $L$-th intensity level. Therefore, two Minkowski-based features and entropy form the feature vector of the proposed metric. Support vector regression (SVR) is used to map these three features to the mean opinion scores (MOS). For the purpose of contrast distortion classification, support vector classifier (SVC) is used to assign a label to each image which indicates to the type of the contrast distortion.

\section{Experimental results}
\label{results}

\subsection{Contrast distorted datasets}

In the experiments, contrast distorted images of four standard datasets are used. The TID2013 \cite{TID2013} dataset contains 125 global contrast changed images, and 125 images with mean shift distortion. CSIQ \cite{MAD} is another dataset that contains 116 global contrast distorted images in total. CCID2014 is a dedicated dataset of contrast distorted images \cite{cyber2015}. It contains 655 contrast distorted images of five types. Gamma transfer, convex and concave arcs, cubic and logistic functions, mean shifting, and a compound function are used to generate these five types of distortions. Please refer to ref. \cite{cyber2015} for detailed explanation. Finally, SIQAD \cite{SIQAD} is used in this paper which contains 140 contrast distorted screen content images. The TID2008 \cite{TID2008} and CID2013 \cite{ICIP2013} datasets are not used in this paper because they are subsets of TID2013 and CCID2014, respectively.

\begin{table*}[htb]
\centering
\scriptsize
\caption{PERFORMANCE COMPARISON OF THE PROPOSED NR-IQA MODEL MDM AND THIRTEEN POPULAR/COMPETING INDICES ON FOUR BENCHMARK DATASETS OF CONTRAST DISTORTED IMAGES}
\label{results1}
\begin{tabular}{|c|c|cccccc|ccc|ccccc|}
\hline
\multicolumn{2}{|c|}{Index}                                                                     & PSNR   & SSIM   & VIF          & IWSSIM       & FSIM$_c$  & MDSI          & RIQMC        & QMC          & RCIQM & QAC    & NIQE   & NSS          & NIQMC & MDM         \\
\multicolumn{2}{|c|}{Type}                                                                      & FR     & FR     & FR           & FR           & FR     & FR           & RR           & RR           & RR      & NR     & NR     & NR           & NR           & NR           \\ \hline
\multirow{2}{*}{\begin{tabular}[c]{@{}c@{}}TID\\ 2013\end{tabular}}  & PCC                      & 0.4755 & 0.5735 & {\bf 0.8458} & 0.6919       & 0.6468 & 0.7028       & 0.8619 & 0.7710       & {\bf 0.8866}  & 0.1683 & -0.0734 & 0.5317       & 0.7225 & {\bf 0.9279} \\
                                                                     & SRC                      & 0.5020 & 0.4992 & {\bf 0.7716} & 0.4528       & 0.4398 & 0.4859       & 0.8010 & 0.7071       & {\bf 0.8541}  & 0.0278 & -0.0652 & 0.4053       & 0.6458 & {\bf 0.8980} \\ \hline
\multirow{2}{*}{CSIQ}                                                & PCC                      & 0.8888 & 0.7891 & 0.9439       & {\bf 0.9614} & 0.9452 & 0.9580       & 0.9605       & 0.9622 & {\bf 0.9645}  & 0.3737 & 0.3025 & 0.8265       & 0.8747 & {\bf 0.9663} \\
                                                                     & SRC                      & 0.8621 & 0.7922 & 0.9345       & {\bf 0.9539}       & 0.9438 & 0.9446 & 0.9501       & 0.9554 & {\bf 0.9569}  & 0.2533 & 0.2284 & 0.7994       & 0.8533 & {\bf 0.9476} \\ \hline
\multirow{2}{*}{\begin{tabular}[c]{@{}c@{}}CCID\\ 2014\end{tabular}} & PCC                      & 0.4112 & 0.8308 & {\bf 0.8588}       & 0.8353       & 0.8204 & 0.8576       & 0.8701 & {\bf 0.8960} & 0.8845  & -0.2765 & 0.4458 & 0.7878 & 0.8438       & {\bf 0.8719} \\
                                                                     & \multicolumn{1}{l|}{SRC} & 0.6743 & 0.8174 & {\bf 0.8349}       & 0.7822       & 0.7657 & 0.8128       & 0.8430 & {\bf 0.8722} & 0.8565  & -0.1419 & 0.3655 & 0.7753 & 0.8113       & {\bf 0.8368} \\ \hline
\multirow{2}{*}{SIQAD}                                                & PCC                      & 0.7529 & 0.8106 & 0.7083       & \textbf{0.8411}       & 0.8204 & 0.7592 & \textbf{0.3697} & 0.2610 & N/A  & 0.1200 & 0.0244 & 0.5770 & 0.2735       & \textbf{0.7890} \\
                                                                     & \multicolumn{1}{l|}{SRC} & 0.6828 & 0.7251 & 0.5269       & \textbf{0.7540}       & 0.7057 & 0.5893       & \textbf{0.3002} & 0.0019 & N/A  & 0.0745 & 0.0717 & 0.2983 & 0.2695       & \textbf{0.6557}  \\ \hline                                                                                                                                          
                                                                     
\end{tabular}
\end{table*}

\subsection{Objective evaluation}

For objective evaluation, two evaluation metrics were used in the experiments: the Spearman Rank-order Correlation coefficient (SRC), and the Pearson linear Correlation Coefficient (PCC) after a nonlinear regression analysis. The SRC and PCC metrics measure prediction monotonicity and prediction accuracy, respectively. The reported PCC values in this paper are computed after mapping quality scores to MOS based on the following logistic function:

\begin{equation}
  \ f(x) = \beta_1\Big(\frac{1}{2}-\frac{1}{1+e^{\beta_2(x-\beta_3)}}\Big)+\beta_4x+\beta_5
  \label{equ:REG}
\end{equation}                    
where $\beta_1$, $\beta_2$, $\beta_3$, $\beta_4$ and $\beta_5$ are fitting parameters computed by minimizing the mean square error between quality predictions $x$ and subjective scores MOS.

Six FR-IQAs including the PSNR, SSIM \cite{SSIM}, VIF \cite{VIF}, IWSSIM \cite{IWSSIM}, FSIM$_c$ \cite{FSIM}, MDSI \cite{MDSI}, and three RR-IQAs, e.g. RIQMC \cite{cyber2015}, QMC \cite{QMC} and RCIQM \cite{RCIQM2017}, and four NR-IQAs including QAC \cite{QAC}, NIQE \cite{NIQE}, NSS \cite{SPL2014} and NIQMC \cite{NIQMC2017} were used in the experiments.

Table \ref{results1} provides a performance comparison between proposed NR-IQA, e.g. MDM, and thirteen FR/RR/NR-IQAs in terms of SRC and PCC. The best performing FR/RR/NR IQAs are highlighted for each category. It can be seen that RR-IQAs that are designated to assess contrast distorted images provide relatively good prediction accuracy on the first three datasets which contain natural images. Among FR-IQAs, the performance of VIF is noticeable for natural images. With a comparison between NR-IQAs, the following conclusions can be drawn. First, the proposed index MDM performs very well on the first three datasets. MDM outperforms NR-IQAs on the four datasets. The recently proposed NR metric for contrast distorted images NIQMC is only comparable to the proposed metric on the CCID2014 dataset. MDM outperforms all the indices listed in Table \ref{results1} on 250 contrast distorted images of the TID2013 dataset. On the other datasets, the proposed index MDM is comparable to the best performing RR metrics RIQMC, QMC and RCIQM. For the screen content dataset SIQAD, only FR-IQA metric IWSSIM has better performance than the proposed metric. However, the popular NR-IQA model NSS \cite{SPL2014} shows inconsistent predictions on different datasets. Also, multi-purpose NR-IQAs like QAC and NIQE have major difficulty in quality assessment of contrast distorted images. It can be concluded that three features of the proposed method are more powerful than the five features of NSS for the purpose of assessing quality of contrast distorted images.

\begin{table}[htb]
\centering
\scriptsize
\caption{Performance comparison of the proposed metric (MDM) and NSS for different train-test setups on the three datasets.}
\label{results2}
\begin{tabular}{c|cc|cc|cc}
\hline
\multirow{2}{*}{NR index} & \multicolumn{2}{c|}{TID2013} & \multicolumn{2}{c|}{CSIQ} & \multicolumn{2}{c}{CCID2014} \\
                          & SRC           & PCC          & SRC         & PCC         & SRC           & PCC          \\ \hline
20\%-80\%                 &               &              &             &             &               &              \\
NSS                       & 0.2507        & 0.3239       & 0.7347      & 0.7491      & 0.7686        & 0.7525       \\
MDM                       & {\bf 0.8702}        & {\bf 0.9091}       & {\bf 0.9243}      & {\bf 0.9332}      & {\bf 0.8214}        & {\bf 0.8564}       \\ \hline
50\%-50\%                 &               &              &             &             &               &              \\
NSS                       & 0.3514        & 0.4702       & 0.7737      & 0.7884      & 0.7807        & 0.7663       \\
MDM                       & {\bf 0.8807}        & {\bf 0.9176}       & {\bf 0.9348}      & {\bf 0.9475}      & {\bf 0.8274}        & {\bf 0.8620}       \\ \hline
80\%-20\%                 &               &              &             &             &               &              \\
NSS                       & 0.4053        & 0.5317       & 0.7994      & 0.8265      & 0.7878        & 0.7753       \\
MDM                       & {\bf 0.8980}        & {\bf 0.9279}       & {\bf 0.9476}      & {\bf 0.9663}      & {\bf 0.8368}        & {\bf 0.8719}       \\ \hline
\end{tabular}
\end{table}

In Table \ref{results2}, the performance of the NR metric NSS \cite{SPL2014} and the proposed metric are listed for different train and test setups. Each dataset is divided into different randomly chosen subsets and the results are reported on the basis of the median value of 1000 times train-test for three cases: 20\% train 80\% test, 50\% train 50\% test, and 80\% train 20\% test. The splits are done in a way that image contents are different for train and test. Hence, for CCID2014 and TID2013 datasets, 0.5333\% train 0.4667\% test and 52\% train 48\% test is used respectively instead of the 50\%-50\% train-test. From the results of the Table \ref{results2}, it can be seen that the proposed metric performs very well with small number of training data. Also, the proposed metric with three features outperforms the five-features metric NSS on the three datasets.

As suggested in \cite{VQEG2003, statistical2006}, we use F-test to decide whether a NR-IQA metric is statistically superior to another NR-IQA index. The F-test is based on the residuals between the quality scores given by an IQA model after applying nonlinear mapping of equation (\ref{equ:REG}), and the mean subjective scores MOS. The ratio of variance between residual errors of an IQA model to another model at 95\% significance level is used by F-test. According to the F-test, the proposed metric is significantly better than the other NR-IQA metrics tested in this paper on different datasets.

\subsection{Contrast distortion classification}

While no-reference image quality assessment of contrast distorted images is of great interest, the classification of contrast distortion types provides very useful additional information that can be used for automatic and fast contrast enhancement. In this paper, the three features of the proposed metric are used to classify type of contrast distortions. The only dataset with more than one type of contrast distortion and with known labels for each distortion type is TID2013 \cite{TID2013}. TID2013 contains 125 distorted images with global contrast change and 125 distorted images with mean shift. Table \ref{accuracy} lists accuracy results of the proposed method and NSS for contrast distortion classification on the TID2013 dataset. In this experiment, image contents for train and test has no overlap. The three features of the proposed method can fairly classify contrast distortions even with small number of training data. However, five features of NSS do not have enough discriminative power to be used for this classification task. Results of Table \ref{accuracy} verify these statements.

\begin{table}[htb]
\centering
\caption{Contrast distortion classification accuracy of the three features of the proposed method and five features of NSS for different setups of train and test.}
\label{accuracy}
\begin{tabular}{cccc}
\hline
NR index & 20\%-80\%       & 50\%-50\%       & 80\%-20\%       \\ \hline
NSS      & 0.5925          & 0.6250          & 0.6400          \\
MDM      & \textbf{0.8650} & \textbf{0.9167} & \textbf{0.9400} \\ \hline
\end{tabular}
\end{table}

While the proposed method for contrast assessment and classification is trained on a fair number of the samples, it is interesting to train the model on larger training sets in order to deal with the different image content. In \cite{APT, ASIQE}, this strategy is used and very promising results are reported. The proposed method will be improved by this strategy in future works. We also observed that the proposed quality assessment metric has limitation in quality assessment of the contrast distorted screen content images. Still, it shows much better performance than the other NR-IQA metrics on the screen content dataset SIQAD. The proposed method is basically more suitable for quality assessment of contrast distorted natural images.

\subsection{Parameters}

The proposed index MDM has two parameters to set, e.g. $q$ and $\rho$. Experimentally, we found that MDM has its maximum prediction accuracy for $q=\{8, 10\}$ and some $50 \leq \rho \leq 130$. Further increasing the value of $q$ will have little effect on the performance. Apart from the performance, being a power of 2 was another consideration on the choice of the parameters $\rho$ and $q$ because MDM runs faster in this case (please refer to subsection Complexity). Therefore, we limited the search domain of the parameters to $q=\{2, 4, 8, 16\}$ and $\rho=\{2, 4, 8, 16, 32, 64, 128\}$. While $\rho=128$ is a slightly better choice than $\rho=64$, $\rho=64$ is used by MDM to maintain numerical stability. The TID2013 dataset was used to chose the best combination of the parameters.

\subsection{Complexity}

To show the efficiency of the proposed metric, a run-time comparison between fourteen IQAs is performed and shown in Table \ref{MDMtime}. The experiments were performed on a Core i7 3.40 GHz CPU with 16 GB of RAM. The IQA model was implemented in MATLAB 2013b running on Windows 7. It can be seen that PSNR and MDM are the top two fastest indices for images with different resolution. Depending on the image resolution, MDM might run faster than PSNR because the code is optimized to calculate power operations in O($\log\rho$) and O($\log q$) instead of O($\rho$) and O($q$) respectively, and that the distorted image is downsampled by a factor of $M = max(2, [min(h, w)/512])$. Here, $h$ and $w$ are image height and width, and $[.]$ is the round operator. In addition, the proposed method only processes the distorted image, while PSNR processes both reference and its distorted version. In comparison with the most competing NR metric NIQMC which is also proposed for contrast distortion assessment, the proposed method is about 180 to 550 times faster. Clearly, the proposed index is highly efficient and can be used in real-time applications.

\begin{table}[htb]
\centering
\scriptsize
\caption{RUN TIME COMPARISON OF IQA MODELS IN TERMS OF MILLISECONDS}
\label{MDMtime}
\begin{tabular}{lccc}
\hline
IQA model & 384$\times$512 & 1080$\times$1920 & 2160$\times$3840 \\ \hline
PSNR      & 5.61    & 37.51     & 145.83    \\
SSIM \cite{SSIM}      & 14.99   & 77.59     & 287.92    \\
VIF \cite{VIF}      & 572.93  & 6162.70   & 25381.32  \\
IWSSIM \cite{IWSSIM}   & 228.11  & 2499.94   & 10471.56  \\
FSIM$_c$ \cite{FSIM}    & 142.06  & 600.23    & 1562.81   \\
MDSI  \cite{MDSI}     & 12.77   & 153.47    & 781.41    \\
RIQMC  \cite{cyber2015}    & 743.90  & 2868.79   & 6313.60   \\
QMC    \cite{QMC}     & 9.40   & 51.38     & 232.42    \\
RCIQM  \cite{RCIQM2017}    & N/A     & N/A       & N/A       \\
QAC    \cite{QAC}    & 151.88  & 1706.15   & 7180.60   \\
NIQE   \cite{NIQE}     & 187.80  & 1726.97   & 6878.10   \\
NSS    \cite{SPL2014}     & 23.92   & 247.13    & 976.11    \\
NIQMC  \cite{NIQMC2017}    & 2897.52 & 10580.00  & 34190.62  \\
$\triangleright$ MDM       & 5.23    & 56.91     & 60.40    \\ \hline
\end{tabular}
\end{table}


\section{Conclusion}
\label{conclusion}
In this paper, an image quality assessment and classification model for contrast distorted images was proposed. The proposed index is very simple and runs in real-time. The proposed index (MDM) uses two Minkowski distance based features and entropy information to assess simple and complex types of contrast distortions. For the first time, the features of the proposed metric were used to classify type of the contrast distortions with a high accuracy. A comparison with the state-of-the-art no-reference IQAs verifies that the proposed metric MDM runs much faster and provides better prediction accuracy on different benchmark datasets than existing NR metrics. In addition, compared to the existing state-of-the-art full reference and reduced reference IQAs, the proposed index shows comparable or better prediction accuracy.

\section*{Acknowledgments}
The authors thank the NSERC of Canada for their financial support under Grants RGPDD 451272-13 and RGPIN 138344-14.


\bibliographystyle{IEEEtran}
\bibliography{egbib2}   

%






\end{document}